\documentclass{article}

\usepackage{adjustbox}

\usepackage[preprint]{neurips_2025}

\usepackage[utf8]{inputenc} 
\usepackage[T1]{fontenc}    
\usepackage{hyperref}       
\usepackage{url}            
\usepackage{booktabs}       
\usepackage{amsfonts}       
\usepackage{nicefrac}       
\usepackage{microtype}      
\usepackage{xcolor}         
\usepackage{graphicx}
\usepackage{array}
\newcolumntype{C}[1]{>{\centering\arraybackslash}p{#1}}
\usepackage{tabularx}
\usepackage{multirow}
\usepackage{booktabs}
\usepackage{subfig}
\usepackage{pgfplots}
\usepackage{pgfplotstable}
\pgfplotsset{compat=newest}
\usepackage{bm}
\usepackage{wrapfig}
\usepackage{lipsum}  

\usepackage{algorithm}
\usepackage[noend]{algpseudocode}

\newtheorem{myproblem}{\textbf{Problem}}
\newtheorem{mydefinition}{\textbf{Definition}}

\title{
\mnameb{}: \underline{C}onstraint-Aware \underline{O}ne-Step \underline{Re}inforcement Learning for Simulation-Guided 
Neural Network Accelerator Design

}

\author{
Yifeng Xiao\thanks{University of California, Berkeley. \texttt{\{yifengx, pierluigi.nuzzo\}@berkeley.edu}} \and
Yurong Xu \thanks{Futurewei Technologies. \texttt{\{yan.ningyan, masood.mortazavi\}@futurewei.com}}
\and Ning Yan\footnotemark[2] \and Masood Mortazavi\footnotemark[2     ]
\and Pierluigi Nuzzo\footnotemark[1]
}

\usepackage{glossaries}

\newacronym{nn}{NN}{neural network}
\newacronym{dnn}{DNN}{deep neural network}
\newacronym{pe}{PE}{processing element}
\newacronym{mac}{MAC}{multiply accumulate unit}
\newacronym{l1}{L1}{local scratchpad}
\newacronym{l2}{L2}{global scratchpad}
\newacronym{noc}{NoC}{networks-on-chip}
\newacronym{hw}{HW}{hardware}
\newacronym{fpga}{FPGA}{field programmable gate arrays}
\newacronym{asic}{ASIC}{application-specific integrated circuit}
\newacronym{rl}{RL}{reinforcement learning}
\newacronym{ga}{GA}{genetic algorithm}
\newacronym{mdp}{MDP}{Markov decision process}
\newacronym{ssmdp}{SSMDP}{single-step Markov decision process}
\newacronym{pdf}{PDF}{probability density function}
\newacronym{kl}{KL}{Kullback-Leibler}
\newacronym{lpa}{LAS}{latency-area-summation}
\newacronym{dse}{DSE}{design space exploration}
\newcommand{\mname}{$\mathsf{CORE}$}
\newcommand{\mnameb}{$\boldsymbol{\mathsf{CORE}}$}

\begin{document}

\maketitle

\begin{abstract}
    Simulation-based design space exploration (DSE) aims to efficiently optimize high-dimensional structured designs under complex constraints and expensive evaluation costs. 
    Existing approaches, including heuristic and multi-step reinforcement learning (RL) methods, struggle to balance sampling efficiency and constraint satisfaction due to sparse, delayed feedback, and large hybrid action spaces. 
    In this paper, we introduce \mname{}, a constraint-aware, one-step RL method for simulation-guided DSE. In \mname{}, the policy agent learns to sample design configurations by defining a structured distribution over them, incorporating dependencies via a scaling-graph-based decoder, and by reward shaping to penalize invalid designs based on the feedback obtained from simulation. 
    {\mname{} updates the policy using a surrogate objective that compares the rewards of  designs within a sampled batch, without learning a value function. This critic-free formulation enables efficient learning by encouraging the selection of higher-reward designs.}    
    We instantiate \mname{} for hardware-mapping co-design of neural network accelerators, demonstrating that it significantly improves sample efficiency and achieves better accelerator configurations compared to state-of-the-art baselines. 
    Our approach is general and applicable to a broad class of discrete-continuous constrained design problems.
\end{abstract}
\section{Introduction}
\label{sec:intro}

Simulation-based \gls{dse} plays a critical role in automated hardware-software co-design, compiler tuning, and system-level optimization. Such optimization tasks typically involve complex, hybrid discrete-continuous design spaces, expensive evaluations through black-box simulations, and strict design constraints. Traditional approaches, such as genetic algorithms~\cite{holland1992genetic} and Bayesian optimization~\cite{lei2021bayesian}, face significant challenges due to sparse and delayed feedback, scalability issues, and limited mechanisms for enforcing structural constraints.

Methods based on \gls{rl}~\cite{kaelbling1996reinforcement, schulman2017proximal} typically frame \gls{dse} as a sequential \gls{mdp} problem~\cite{krishnan2023archgym, jiang2021delving}, requiring long rollout-based exploration or value function approximations that are often impractical for expensive simulation environments. Furthermore, these methods frequently rely on heuristic masking and coarse discretizations, reducing exploration efficiency and potentially violating feasibility constraints during training. Thus, an efficient method explicitly designed for structured, constraint-aware sampling remains an important open challenge.

{In this paper, we propose \mname{} (\underline{C}onstraint-aware \underline{O}ne-step \underline{RE}inforcement learning), a one-step \gls{rl} method~\cite{ghraieb2021single} specifically tailored for simulation-guided structured \gls{dse}. 
\mname{} generates complete candidate configurations in a single step by learning a structured distribution over design variables, 
avoiding the inefficiencies of sequential rollout-based methods, the need to maintain intermediate design states, and the reward sparsity issues that arise in multi-step \gls{rl}.
To ensure constraint satisfaction and exploit parameter dependencies, \mname{} introduces a novel scaling-graph-based decoder that enforces parameter dependencies during sampling and a reward shaping mechanism to penalize invalid configurations.
The policy is updated by a critic-free surrogate objective based on batch-relative reward~\cite{mortazavi2022, jiayu2024, guo2025deepseek}, which are obtained efficiently through parallel simulation of sampled designs, eliminating the need for value functions.}

We demonstrate \mname{}'s effectiveness in a challenging application domain: the co-design of hardware and mapping strategies for spatial \gls{dnn} accelerators~\cite{chen2016eyeriss,nvdla2017,du2015shidiannao} to accelerate \gls{dnn} inference by leveraging parallelism and data reuse.
Given a \gls{dnn} load, the accelerator architecture is designed to execute a fixed-size tensor computation while the \textit{mapping strategy} determines how the computation is distributed among the \textit{hardware resources} to achieve the best performance.
This setting presents a representative and challenging testbed due to its rich structure, large combinatorial space, and costly simulation feedback, making it well-suited to assess the strengths of our method.
We focus exclusively on this domain to enable a thorough evaluation, but the methodology is designed to generalize to other structured, simulation-based optimization problems.

Our contributions can be summarized as follows:
\begin{itemize} 
    \item We propose \mname{}, a critic-free, one-step \gls{rl} framework integrating parallel evaluations for simulation-based structured \gls{dse} problems.
    \item We introduce a scaling-graph-based decoding strategy that improves the generation of feasible configurations by explicitly modeling parameter dependencies and constraints.
    \item We design a constraint-aware reward-shaping mechanism to penalize invalid configurations, significantly enhancing the exploration efficiency.
    \item We evaluate \mname{} on the co-design of hardware and mapping strategies for DNN accelerators, showing that our method can achieve at least 15$\times$ improvement in both latency and latency-area-sum metrics with fewer sample designs, compared to state-of-the-art methods.
\end{itemize}

The rest of the paper is organized as follows. 
Section~\ref{sec:prelim} formulates the problem and introduces the background on \gls{dse} and \gls{dnn} accelerator design.
Section~\ref{sec:algo} presents the proposed optimization algorithm while Section~\ref{sec:results} reports the evaluation results. Section~\ref{sec:conclusion} concludes the paper.

\section{Background and Problem Formulation}
\label{sec:prelim}

\paragraph{Simulation-Based Design Space Exploration.}

Simulation-based \gls{dse} methods have evolved from traditional heuristic-based algorithms, such as genetic algorithms~\cite{holland1992genetic} and simulated annealing~\cite{bertsimas1993simulated}, to more structured optimization techniques, including surrogate modeling and Bayesian optimization~\cite{lei2021bayesian}. These problems are generally not amenable to analytical solutions due to the complexity and non-differentiable nature of the simulation-based objective metrics.
We formalize the design problem:

{\begin{myproblem}
Given a design space $\mathcal{D}$, a simulator $U: \mathcal{D} \rightarrow \mathbb{R}^J$ that returns $J$ performance metrics for a design configuration $\xi \in \mathcal{D}$, a set of parameter dependency constraints $\{g_i(\xi) \le 0\}_{i=1}^K$, a set of performance constraints $\{h_j(U(\xi)) \le 0\}_{j=1}^M$, and a scalar reward function $R: \mathbb{R}^J \rightarrow \mathbb{R}$, the goal is to find an optimal configuration $\xi^*$ that solves:
\begin{align}
    \underset{\xi \in \mathcal{D}}{\text{\rm maximize}} \quad R(U(\xi))
    \quad \text{\rm s.t.} \quad
    \begin{cases}
        g_i(\xi) \le 0, & \forall i \in \{1, \dots, K\} \\
        h_j(U(\xi)) \le 0, & \forall j \in \{1, \dots, M\}
    \end{cases}
    \label{eq:opt}
\end{align}
\end{myproblem}}

\paragraph{\gls{rl}-Based Design Space Exploration.}

Recent work leverages \gls{rl} to explore complex design spaces in domains such as memory controller tuning, system-on-chip  design~\cite{krishnan2023archgym}, and 
microarchitecture search~\cite{bai2024towards}. These methods typically model the task as a multi-step \gls{mdp}, where design choices are made sequentially and rewards are obtained only after a full design evaluation. This leads to sparse, delayed feedback and requires maintaining partial design states or heuristic masking~\cite{jiang2021delving} to enforce constraints, complicating learning and scaling.
We instead adopt a one-step \gls{rl} formulation~\cite{ghraieb2021single}, where the policy generates complete design candidates in one step, enabling efficient parallel sampling and eliminating the need for sequential rollouts, intermediate state designs, and sparse reward propagation. 

\begin{figure}[tb!]
    \centering
    \subfloat []
    {\adjustbox{valign=c}{\includegraphics[width=0.2\columnwidth]{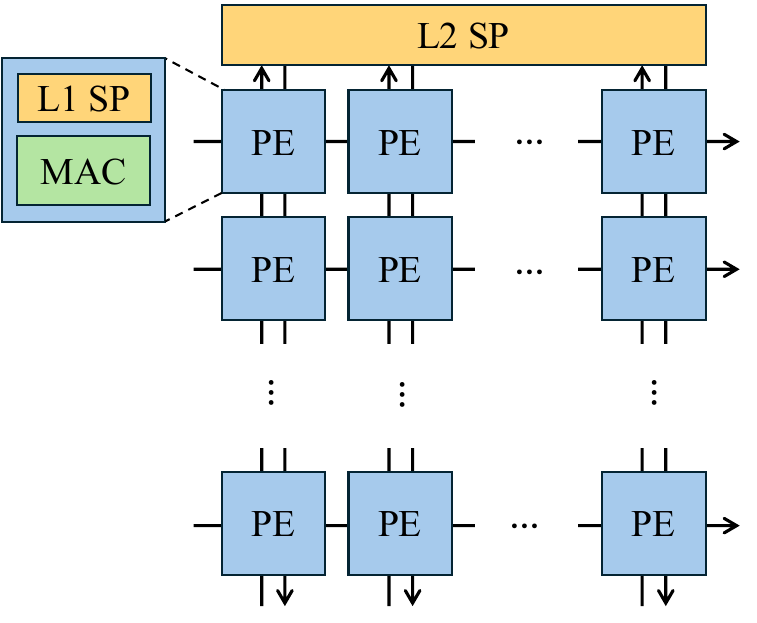}} \label{fig:hw}} 
    \subfloat []
    {\adjustbox{valign=c}{\includegraphics[width=0.3\columnwidth]{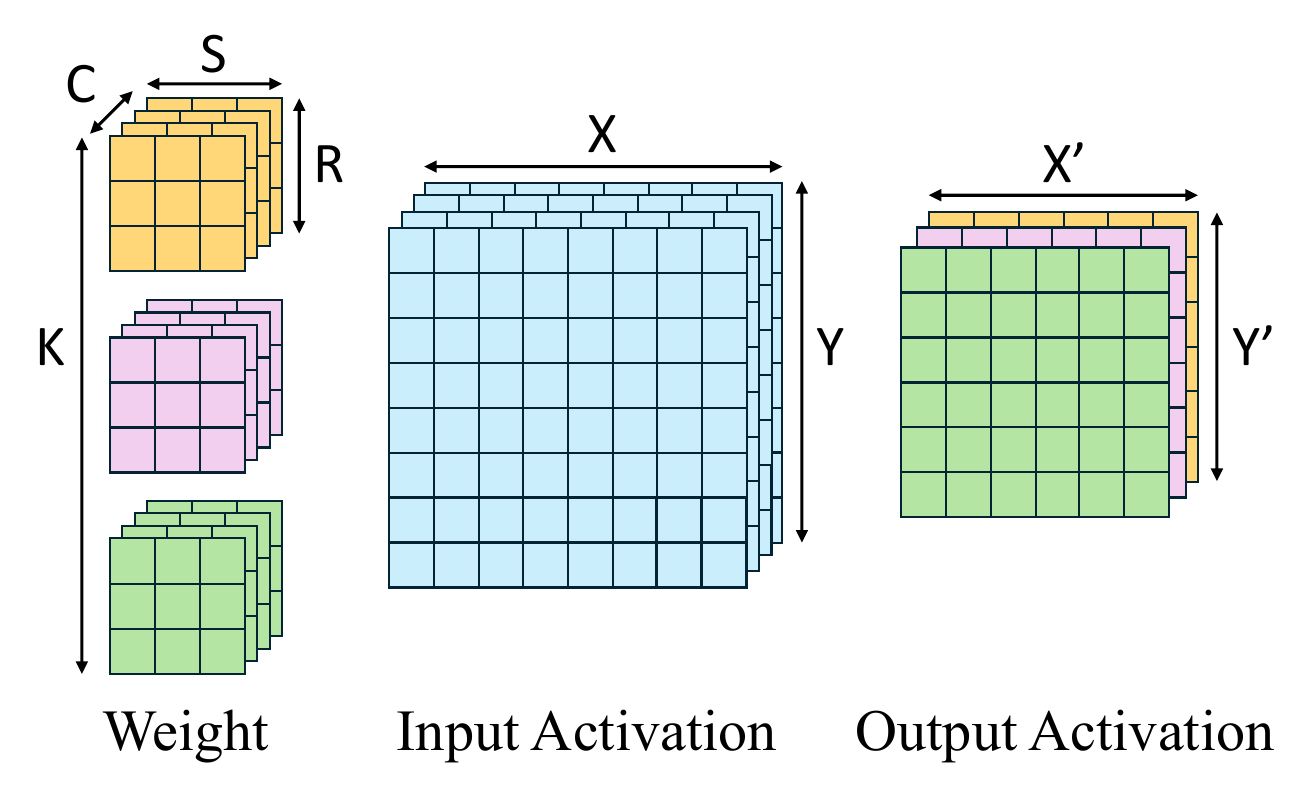}} \label{fig:conv}}
	\subfloat[]
    {\adjustbox{valign=c}{
        \setlength\tabcolsep{0.5em}
		\renewcommand{\arraystretch}{1.1}
		\resizebox{0.46\columnwidth}{!}{\begin{tabular}{cc|c}
			\toprule
			\multicolumn{2}{c|}{Parameter}                                                         & Value                   \\ \hline
			\multicolumn{1}{c|}{\multirow{4}{*}{Hardware resources}}                     & \# of PEs  & $2:1024:2$ \\
			\multicolumn{1}{c|}{}                                                 & L2 buffer size (bytes) & $1:2^{32}$ \\
			\multicolumn{1}{c|}{}                                                 & L1 buffer size (bytes) & $1:2^{32}$                  \\
			\hline
			\multicolumn{1}{c|}{\multirow{9}{*}{{\begin{tabular}{c}Mapping strategies \\ of level $i$ in one layer \end{tabular}}}} & Loop order     & $<\mathsf{S_i,R_i,K_i,C_i,X_i,Y_i}>$                    \\
			\multicolumn{1}{c|}{}                                                 & Parallelization dimension $\mathsf{P}_i$ &  $\mathsf{S_i,R_i,K_i,C_i,X_i,Y_i}$                      \\
			\multicolumn{1}{c|}{}                                                 & Level of parallelism ${P_i}$ & $1:\mathsf{P}_i$               \\
			\multicolumn{1}{c|}{}                                                 & ${S}_i$              & $1:\mathsf{S}_i$                \\
			\multicolumn{1}{c|}{}                                                 & ${R}_i$              & $1:\mathsf{R}_i$                \\
			\multicolumn{1}{c|}{}                                                 & ${K}_i$              & $1:\mathsf{K}_i$                \\
			\multicolumn{1}{c|}{}                                                 & ${C}_i$              & $1:\mathsf{C}_i$                \\
			\multicolumn{1}{c|}{}                                                 & ${X}_i$              & $1:\mathsf{X}_i$                \\
			\multicolumn{1}{c|}{}                                                 & ${Y}_i$              & $1:\mathsf{Y}_i$                \\ \bottomrule
			\end{tabular}}
    }\label{tab:space}}
    \caption{
		(a) Hardware resources for a 2-level spatial \gls{dnn} accelerator; (b) Tensor dimensions for convolutional layers. 
        (c) Design space of hardware resources and mapping strategies. 
        }
    \label{fig:design_table_pipeline}
\end{figure}

{\paragraph{Spatial \gls{dnn} Accelerator Design.}  
We instantiate Problem~\ref{eq:opt} for the co-design of hardware resources and mapping strategies in spatial \gls{dnn} accelerators. 
Each design configuration $\xi \in \mathcal{D}$ consists of a structured combination of hardware parameters and mapping strategies.
The simulator $U(\xi)$ evaluates a given design based on performance metrics such as latency, area, and power, which are encoded into a scalar reward via $R(U(\xi))$. 
The constraint functions \( g_i(\xi) \) enforce dependencies over parameters (e.g., buffer bounds, tile hierarchy), while \( h_j(U(\xi)) \) encode system-level performance constraints such as area budgets across target platforms.}

Prior work typically searches for efficient mappings on fixed hardware or tunes hardware configurations under a fixed mapping strategy~\cite{kao2020confuciux,kao2020gamma,krishnan2023archgym,zheng2020flextensor}.
While hardware-mapping interdependencies suggest that joint optimization can yield better performance, this remains challenging due to the vast combined design space~\cite{digamma,kao2020gamma}. 
Recent approaches to address this problem include heuristic search~\cite{digamma} and two-step optimization~\cite{xiao2021hasco, rashidi2023unico}. However,  these methods are limited by sampling inefficiency and scalability for large design spaces.
In contrast, \mname{} performs joint optimization over the design space by embedding structured design parameters into a continuous distribution. 
A scaling-graph-based decoder enforces constraints and dependencies during sampling
enabling efficient and scalable exploration of valid configurations.

\paragraph{Co-Design Space and Parameter Dependencies.}\label{sec:space}
As shown in Table~\ref{tab:space}, the hardware design space includes the number of \glspl{pe}, and the sizes of L1 and L2 buffers. The mapping design space for each memory level $i$ includes the loop ordering, tile sizes ($S_i, R_i, K_i, C_i, X_i, Y_i$), the parallelization dimension $\mathsf{P}_i$, and the level of parallelism $P_i$. Here, the subscript $i$ denotes the memory hierarchy level (e.g., $i = 1$ for L1, $i = 2$ for L2).
Further details on parameter ranges and encoding are provided in Appendix~\ref{sec:dnn}.
The design space exhibits rich structural dependencies. For instance, buffer sizes and \glspl{pe} constrain the feasible L2 buffer size; tile sizes must satisfy \( D_i \le D_{i+1} \) across memory levels to preserve hierarchical partitioning~\cite{kao2020gamma}, where each $ D_i \in \{S_i, \ldots, Y_i\} $ denotes the tile size of a particular loop dimension at memory level $ i $.; and the number of \glspl{pe} bounds the degree of parallelism via the tile size in the selected parallel dimension.

\section{\mname{} Framework}
\label{sec:algo}

\begin{figure}[t]
	\centering
	\includegraphics[width=0.88\columnwidth]{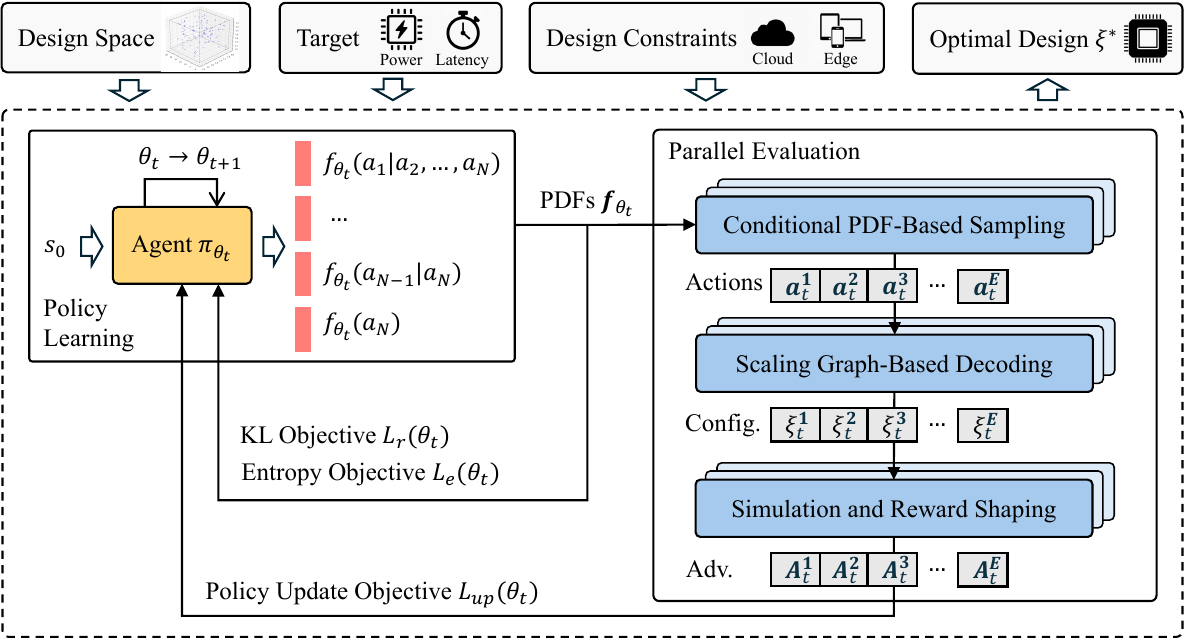}
    \vspace{2mm}
	\caption{Overview of the \mname{} framework for simulation-based design space exploration. 
    }
    \label{fig:ssrl}
\end{figure}

In this section, we introduce the \mname{} framework (Fig.~\ref{fig:ssrl}) for simulation-based \gls{dse}. \mname{} employs an one-step \gls{rl} formulation, where the policy models conditional \glspl{pdf} over structured design actions.
Unlike traditional \gls{rl} methods, \mname{} decouples policy learning from evaluation via a parallel pipeline consisting of conditional sampling, scaling-graph-based decoding, and simulation-driven reward feedback. This design supports high-throughput exploration under delayed and expensive feedback regimes.
The policy is updated using a surrogate objective defined by the relative advantages of the sampled designs, enabling sample-efficient learning without value functions.
We describe each component of the framework in the following sections.

\subsection{One-Step Markov Decision Process and Sampling Policy}
\label{sec:sample}

\begin{mydefinition}[One-Step \gls{mdp}]
A one-step \gls{mdp} is defined as the tuple $\mathcal{M} = (s_0, \mathcal{A}, {R})$, where $s_0 \in \mathcal{S}$ denotes the single state of the environment,
$\mathcal{A}$ denotes 
the action space, and the reward function $R$ is defined as $R(s_0, {a}): \{s_0\} \times \mathcal{A} \rightarrow \mathbb{R}$, ${a} \in \mathcal{A}$.
The agent interacts with the environment in episodes of length $1$.
\end{mydefinition}

We model the design exploration process as a one-step \gls{mdp}, where a policy network $\pi_\theta$ represents a joint distribution over design actions conditioned on a fixed input state $s_0$. 
Since the environment is stateless, $s_0$ acts as a static context vector and remains unchanged throughout training.
The joint distribution $\pi_\theta(a_1, \ldots, a_N; s_0)$ is factorized as a product of conditional probability density functions:
\begin{align}
    \pi_\theta(a_1, ..., a_N; s_0) \!=\! \prod_{i=1}^N \pi_{i, \theta} \!=\! \prod_{i=1}^{N} f_{i, \theta} (a_i | a_{i+1}...a_N; s_0),
\end{align}
where each $f_{i, \theta}$ is a conditional \gls{pdf} corresponding to an action $a_i$,
so the compound action $\bm{a} = (a_1, \ldots, a_N)$ is sampled from the joint distribution.
This factorization captures statistical dependencies between design actions, allowing the model to learn probabilistic preferences for combinations of actions. 
These learned correlations are distinct from the structural dependencies encoded in the scaling graph decoder (Section~\ref{sec:decode}), which ensures that the sampled actions are mapped to valid, feasible configurations. 
We consider two types of distributions for $f_{i,\theta}$:
\begin{itemize}
    \item The \textit{categorical distribution} models discrete actions drawn from a finite set of categories, parameterized by a probability vector $\mathbf{p}$, where $p_{i}$ represents the probability of selecting the $i$-th category. 
    \item The \textit{Beta distribution} usually models continuous actions on the interval $[0,1]$, parameterized by two parameters, $\alpha$ and $\beta$, which control the shape of the distribution. 
\end{itemize}
We use Beta distributions as continuous relaxations for discrete actions when the discrete space is large or context-dependent. 
A continuous value is sampled from the Beta distribution and rounded into a discrete value, as described in Section~\ref{sec:decode}.
For the design space shown in Table~\ref{tab:space}, 
we sample the level of parallelism $P_i$, which includes $6$ choices, using the categorical distribution. 
The remaining parameters are sampled from beta distributions to reduce the output dimension of the policy \gls{nn} and incorporate structural dependencies.

\subsection{Decoding Actions to Configurations}
\label{sec:decode}

\begin{figure*}[tb!]
	\centering
	\includegraphics[width=0.95\textwidth]{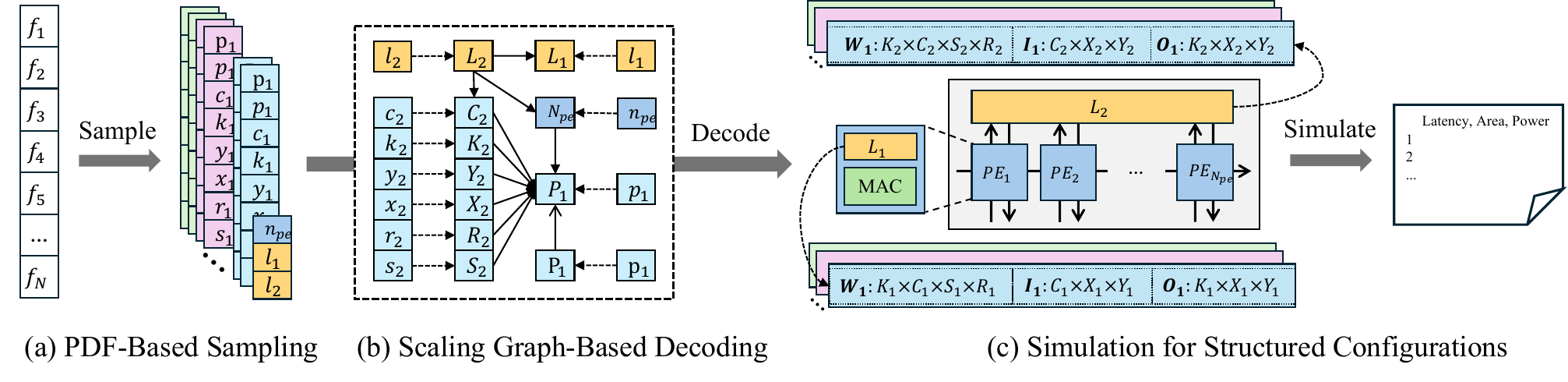}
	\caption{ 
    Evaluation pipeline for hardware-mapping co-design.
    }
    \label{fig:decode}
\end{figure*}

In each training episode, we use $\pi_\theta$ to sample a batch of $E$ compound design actions $\{\bm{a}_k\}_{k=1}^E$, where each $\bm{a}_k$ can be decoded into a design configuration.
We instantiate the evaluation procedure of \gls{dnn} accelerator in Fig.~\ref{fig:decode}, where lowercase letters represent sampled actions, uppercase letters indicate decoded configuration parameters, and color-coded mapping components distinguish design strategies across layers.
We introduce a decoding technique that maps actions from the action space $\mathcal{A}$ to structured design space $\mathcal{D}$, while satisfying parameter dependency constraints.

\paragraph{Action Discretization.}

For {independent discrete parameters} with a wide range of values, we use the Beta distribution to sample an action $b \in [0, 1]$ and then round it into a discrete value $B$.
Generally, assuming a range $[B_{low}, B_{up}]$ and a step size $B_s$, we discretize the action as follows:
\begin{align}
    B = B_{low} + \left\lfloor \left(\frac{B_{up} - B_{low}}{B_s} + 1\right) b \right\rfloor B_s, \label{eq:decode}
\end{align}
where $\lfloor x \rfloor$ denotes the floor function, and the action $b$ is scaled to match the range of the parameter $B$. By taking the number of \glspl{pe} as an example, we have $512$ choices, i.e., $N_{pe}$ can be selected in the range from $2$ to $1024$ with steps of $2$.
Using the beta distribution to produce an action $n_{pe} \in [0, 1]$, we quantize this action as  
$N_{pe} = 2 + 2\lfloor 512  n_{pe} \rfloor$.

\paragraph{Scaling-Graph-Based Decoding.}

\begin{wrapfigure}{r}{0.45\textwidth}
    \centering
    \includegraphics[width=0.42\textwidth]{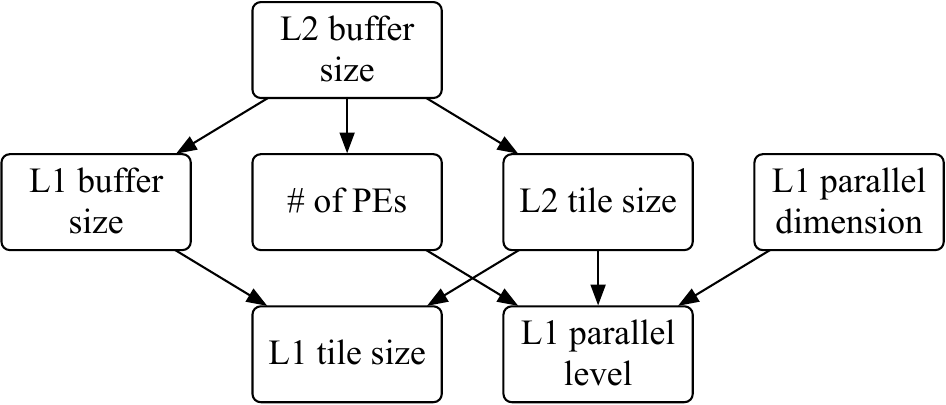}
    \caption{Scaling graph for DNN accelerator.}
    \label{img:scal}
    \vspace{-1em}
\end{wrapfigure}

To effectively navigate structured parameter spaces with interdependencies, we introduce a decoding strategy based on a \emph{scaling graphs}, which dynamically adjusts the parameter bounds, improving feasibility and accelerating convergence during exploration~\cite{eyerman2009mechanistic}.
As shown in Fig.~\ref{img:scal}, a scaling graph captures structural dependencies between design parameters. Each node represents a design variable, and each directed edge encodes a constraint or scaling relationship from a source (influencing) parameter to a target (dependent) parameter. Specifically, the decoded value of the source constrains the feasible range of the target variable, such as the upper bound.

Given a set of source parameters with decoded values $\{ A_i \}$, and a target parameter bounded within $[B_{low}, B_{up}]$ with step size $B_s$, we consider the upper bound relation and a sampled action $b$ such that:
\begin{align}
    B = B_{low} + \left\lfloor \left(\frac{\min_i\{A_i\} - B_{low}}{B_s} + 1\right) b \right \rfloor B_s,
\end{align}
where we replace $B_{up}$ with $\min_i\{A_i\}$ in Equation~\eqref{eq:decode} to ensure that the upper bound is determined by the source nodes.
Intuitively, this scaling ensures that sampled actions respect inter-parameter dependencies by dynamically constraining action ranges according to previously decoded parameters.
Decoding follows the topological order of the scaling graph to respect dependencies.

In the example shown in Fig.~\ref{fig:decode} (b),
the upper bound for the level of parallelism $P_1$ is constrained by the number of \glspl{pe} and the corresponding tile size in level $2$, which is determined by the selected parallelization dimension $\mathsf{P}_1$.
Assuming $P_1 \ge P_{low}$ and the selected parallel dimension is $\mathsf{X}$, i.e., $\mathsf{P}_1 = \mathsf{X}$, we decode $P_1$ as follows:
\begin{align}
P_1 = P_{low} + \lfloor (\min\{N_{pe}, X_2\} - P_{low} + 1) p_1 \rfloor.\end{align}
In summary, the scaling graph enables constraint-aware decoding by dynamically adjusting parameter bounds during sampling. This ensures that the sampled configurations are feasible and accelerates learning.

\subsection{Parallel Sampling Policy Optimization}
\label{sec:pspo}
We customize proximal policy optimization~\cite{schulman2017proximal, hsu2020revisiting} for design exploration. The objective function $R(\xi)$
in Problem~\eqref{eq:opt} is obtained from the simulation results of a design point $\xi$.
We incorporate the constraint-aware reward shaping to formulate a surrogate objective function $L(\theta_t)$, which is used to update the policy \gls{nn}.

\paragraph{Reward and Surrogate Advantage.}
Let $\{\xi_k\}_{k=1}^E$ denote a batch of $E$ design samples, where each $\xi_k \in \mathcal{D}$ is a candidate design point from the design space $\mathcal{D}$. 
Each design is evaluated by a simulator or cost model, which returns $J$ performance metrics (e.g., latency, power, area).  
We define the simulator as a function:
$U : \mathcal{D} \rightarrow \mathbb{R}^J$, which returns a set of metrics value for a design point $\xi$.
Then, the optimization target can be expressed as a weighted sum of the metrics:
\begin{align}
    R(\xi_k) = \bm{w}^\top U(\xi_k), \label{eq:reward}
\end{align}  
where $\bm{w} \in \mathbb{R}^J$ is a user-defined weight vector applied to the simulator outputs.
If a lower value of a metric (e.g., latency or energy consumption) indicates a better design, we assign the corresponding weight $w_j < 0$ so that the scalar reward $R(\xi_k)$ increases as design quality improves. This ensures that all objectives are aligned under a reward maximization framework.

{At each training episode $t$, the agent samples a batch of $E$ design points $\{\xi_k\}_{k=1}^E$ and evaluates them using parallel simulation to obtain a scalar reward $R(\xi_k)$ for each sample. We define the \textit{running reward} $\hat{R}_t$ to track the exponential moving average of batch rewards, which gives higher weight to recent batches while gradually discounting earlier ones:
\begin{align}
    &\hat{R}_t = \alpha_r \mathbb{E}_{\xi \sim \pi_{\theta_t}}[R(\xi)] + (1 - \alpha_r) \hat{R}_{t-1}, \quad \hat{R}_{0} = 0, \label{eq:running}\\
    &\mathbb{E}_{\xi \sim \pi_{\theta_t}}[R(\xi)] \approx \frac{1}{E} \sum_{k=1}^{E} R(\xi_k), \label{eq:appr}
\end{align}
where the expectation over the policy distribution is approximated by the empirical mean of the current batch, and $\alpha_r \in [0, 1]$ is the renewal rate.}
Based on this estimate, we define the \textit{surrogate advantage} $A_t(\xi_k)$ to measure the relative quality of each sampled design in the batch, computed as the difference between its reward and the running average:
\begin{align}
    A_t(\xi_k) = R(\xi_k) - \hat{R}_t.
    \label{eq:adv}
\end{align}

\paragraph{Constraint-Aware Reward Shaping.}

If a design point violates a required constraint $\phi$, we apply a scaling penalty to quantify the degree of violation.
Given a constraint $h(U(\xi_k)) \le 0$, if it is violated, the reward is updated as follows:
\begin{align}
    R(\xi_k) = \bm{w}^\top U(\xi_k) - \alpha_c h(U(\xi_k)), \label{eq:reward2}
\end{align}
where $\alpha_c$ is the violation penalty rate.

If a design point cannot be simulated, possibly due to ignoring certain dependencies or architectural constraints, we call it an \textit{anomalous design}.
We define a penalty to push the reward below the average, discouraging the agent from generating similar samples in the future.
The reward for an anomalous design $\xi'$ in episode $t$ is computed as:
\begin{align}
    R_{t}(\xi') = \begin{cases}
        \min (\mathbb{E}_{\xi \sim \pi_{\theta_{t-1}}}[R(\xi)], ~\hat{R}_{t-1}) - \alpha_p \mathbb{E}_{\xi \sim \pi_{\theta_t}}[R(\xi)], \quad &t > 1 \\
        R_{\text{ano}} \quad &t = 1
    \end{cases}
    \label{eq:reward3}
\end{align}
where $\alpha_p$ is the anomalous design penalty rate, $R_{\text{ano}}$ is the initial reward for the first episode, and the expectation is approximated by the empirical mean of the batch.

\begin{algorithm}[t]
    \caption{{Constraint-Aware One-Step \gls{rl} for Design Exploration}}
    \label{alg:core}
    \small
    \begin{algorithmic}[1]
    \Require Design Space $\mathcal{D}$, Policy \gls{nn} $\pi_\theta$, input state $s_0$, scaling graph $G$, batch size $E$, maximum number of episodes $t_{m}$, target reward $\overline{R}$, dependency constraints $\{g_i(\xi) \le 0\}_{i=1}^K$, performance constraints $\{h_j(U(\xi)) \le 0\}_{j=1}^M$, weights $\bm{w}$ for objective metrics, and learning rate $\eta$.
    \State $ t \gets 1$, $\hat{R}_{0} \gets 0$.
    \State Initialize parameters $\theta_1$ of policy network $\pi_\theta$.
    \While{$t < t_{m}$}
        \State $\bm{a}_{1}, \cdots, \bm{a}_{E} \sim \pi_{\theta_t}(s_0)$; \Comment{Sample $E$ compound actions (Sec.~\ref{sec:sample})}
        \State $\xi_{1}, \cdots, \xi_{E} \gets \mathsf{Decode}(<\bm{a}_{1}, \cdots, \bm{a}_E>, \mathcal{D}, \{g_i\}_{i=1}^K)$; \Comment{Decode via scaling graph (Sec.~\ref{sec:decode})}
        \State $R(\xi_1), \cdots, R(\xi_E) \gets \mathsf{Reward}(<U(\xi_1), \cdots, U(\xi_E)>, \bm{w}, \{h_j\}_{j=1}^M)$; \Comment{Equation~\eqref{eq:reward}, \eqref{eq:reward2}, \eqref{eq:reward3}}
        \State $R_{max}, \xi_{best} \gets \mathsf{FindBest}(R(\xi_1), \cdots, R(\xi_E))$; \Comment{Find $\xi$ with the maximum reward}
        \If{$R_{max} > \overline{R}$}
            \State \textbf{break};
        \EndIf
        \State $\hat{R}_t \gets \alpha_r \frac{1}{E} \sum_{k=1}^{E} R(\xi_k) + (1 - \alpha_r) \hat{R}_{t-1}$; \Comment{Equation~\eqref{eq:running}, \eqref{eq:appr}}
        \State $L(\theta_t) \gets \mathsf{ComputeObj}(<\!R(\xi_1), \cdots, R(\xi_E)\!>, \hat{R}_t, \pi_{\theta_t})$; \Comment{Equation~\eqref{eq:adv}, \eqref{eq:up}, \eqref{eq:kl}, \eqref{eq:ent}}
        \State $\theta_{t+1} \gets \theta_t + \eta \cdot \nabla_\theta L(\theta_t)$;\Comment{Policy update via gradient ascent}
        \State $t \gets t + 1$;
    \EndWhile
    \State \Return $\xi_{best}$;
    \end{algorithmic}
\end{algorithm}

\paragraph{Surrogate Objective.}

The surrogate objective function comprises the conditional update objective, the \gls{kl} objective, and the entropy objective~\cite{mortazavi2022}. 
First, we update the policy parameters $\theta$ based on the surrogate advantage through the conditional update objective, which is computed as follows:
\begin{align}
    L_{up}(\theta_{t}) = \mathbb{E}_{\xi \sim \pi_{\theta_t}}\left[\frac{\pi_{\theta}(a_1, ..., a_N; s_0)}{\pi_{\theta_t}(a_1, ..., a_N; s_0)}   A_t(\xi_k)\right]. \label{eq:up}
\end{align}
Moreover, a \gls{kl}-regularizer is added to regulate the update rate of the policy as follows: 
\begin{align}
    L_r(\theta_{t}) = -\beta_r \sum_{i=1}^N \mathbb{D}_{KL}(\pi_{i, \theta} || \pi_{i, \theta_{t}}), \label{eq:kl}
\end{align}
where $\mathbb{D}_{KL}$ is the forward \gls{kl} divergence~\cite{hsu2020revisiting} and
$\beta_r$ is a factor for the \gls{kl} objective, encouraging the updated policy to stay close to the current policy.
{Lastly, to balance exploration and exploitation, we include an entropy regularization term that encourages the policy to maintain uncertainty in its action distributions. This regularization prevents early convergence to suboptimal, overconfident actions and encourages diverse exploration early in training.
To reduce unnecessary randomness later, we multiply the entropy bonus by a decaying factor $\beta_e$~\cite{mortazavi2022},  which gradually shifts the policy from exploration to exploitation:}
\begin{align}
    L_e(\theta_{t}) = \beta_e \sum_{i=1}^N \mathbb{H}(\pi_{i, \theta}), \label{eq:ent}
\end{align}
where $\mathbb{H}$ is the entropy function.
With the objective function $L(\theta_t) = L_{up}(\theta_t) + L_r(\theta_t) + L_e(\theta_t)$, optimization algorithms such as gradient ascent can be used for back-propagation to update the policy \gls{nn} parameters $\theta$, guiding the model toward an optimal policy. 
Our framework is summarized in Algorithm~\ref{alg:core}. Training terminates when either the maximum number of episodes $t_{\text{m}}$ is reached or the target reward threshold $\overline{R}$ is achieved.
\section{Experiments}\label{sec:results}

\subsection{Experiment Setup}

\paragraph{Policy and Hyperparameters} 

We implement \mname{} in Python using the PyTorch library and evaluate it on a set of \gls{dnn} models for various applications. 
The policy network is a 4-layer multilayer perceptron with ReLU activations (see Appendix~\ref{sec:nn} for architecture details).
It is trained for $2000$ episodes using the Adam optimizer with default beta parameters, a learning rate of $10^{-5}$,
and a batch size of $32$, which corresponds to the number of parallel CPU threads used for simulation. 
The entropy coefficient $\beta_e$ is linearly decayed from $1.0$ to $0.02$, the surrogate reward is computed with a renewal rate $\alpha_r$ of $0.2$, and other rate factors are fixed at $1.0$. 
No hyperparameter tuning is performed.
We implement all experiments on a server with an NVIDIA V100 GPU.

\paragraph{\gls{dnn} Accelerator Setup}

We consider a 2-level mapping strategy for the accelerator design space in the experiments.
We evaluate seven \gls{dnn} models across vision (ResNet-18, ResNet-50, MobileNetV2, VGG-16), language (BERT), and recommendation (DLRM, NCF) domains~\cite{kao2020gamma}.
For transformer models like BERT, 
the mapping space is defined to 
capture matrix multiplications in attention and feed-forward layers. 
This work focuses on inference configuration only.

Area constraints for \glspl{pe} and buffers are imposed to reflect realistic platform settings: $0.2\ mm^2$ for edge devices and $7.0\ mm^2$ for cloud platforms~\cite{shao2019simba}.
While DRAM bandwidth and interconnects vary in practice, we abstract these under a unified simulation interface to maintain comparability.
The simulation-based evaluation uses the MAESTRO cost model~\cite{kwon2020maestro} for performance and Synopsys DC, Cadence Innovus for area estimation (via RTL synthesis with the Nangate 15nm library)~\cite{digamma}. We set a fixed sampling budget of 40,000 for all algorithms to maintain consistency.

\begin{table*}[tb]
  \centering
  \caption{{Log$_{10}$-scaled performance comparison of baseline methods and ablation studies with \mname{} on cloud and edge platforms.}}
  \label{tab:comp}
  \renewcommand{\arraystretch}{1.1}
  \resizebox{\textwidth}{!}{
  \begin{tabular}{c|C{1.1cm}C{1.1cm}C{1.1cm}C{1.1cm}C{1.1cm}|
                   C{1.1cm}C{1.1cm}C{1.1cm}C{1.1cm}C{1.1cm}}
  \toprule
  Objective & \multicolumn{5}{c|}{Latency (log$_{10}$ cycles)} & \multicolumn{5}{c|}{Latency-Area-Sum (log$_{10}$)} \\ \hline
  Method    & GA & HASCO & w/o. rs. & w/o. sc. & \mname{} & GA & HASCO & w/o. rs. & w/o. sc. & \mname{} \\ \hline
  \multicolumn{11}{c}{Cloud  Platform} \\ \hline
Resnet18 & 7.28 & 6.80 & 6.68 & 5.26 & \textbf{4.62} & 7.44 & 6.79 & 6.76 & 6.14 & \textbf{5.66} \\
Resnet50 & 7.29 & 7.30 & 7.31 & 6.47 & \textbf{5.47} & 7.40 & 6.92 & 7.34 & 6.97 & \textbf{6.18} \\
Mbnet-V2 & 7.01 & 6.79 & 6.20 & 5.80 & \textbf{4.33} & 6.89 & 6.75 & 6.30 & 6.54 & \textbf{5.14} \\
BERT     & 7.31 & 6.85 & 6.74 & 5.95 & \textbf{5.60} & 7.08 & 6.74 & 6.94 & 6.52 & \textbf{5.99} \\
NCF      & 3.50 & 3.48 & 3.66 & 3.47 & \textbf{2.17} & 4.65 & \textbf{3.41} & 4.53 & 4.16 & 4.06 \\
DLRM     & 2.57 & 2.60 & 2.83 & 3.73 & \textbf{2.03} & 4.09 & \textbf{3.72} & 4.06 & 4.08 & 4.05 \\
VGG16    & 7.85 & 7.43 & 7.51 & 5.84 & \textbf{5.05} & 7.92 & 7.65 & 7.52 & 7.12 & \textbf{6.08} \\
\hline\multicolumn{11}{c}{Edge Platform} \\ \hline
Resnet18 & 7.20 & 7.06 & 7.03 & 5.33 & \textbf{5.31} & 7.39 & 6.79 & 7.04 & 6.50 & \textbf{5.59} \\
Resnet50 & 7.37 & 7.12 & - & 6.74 & \textbf{5.47} & 7.25 & 6.92 & - & 7.02 & \textbf{6.26} \\
Mbnet-V2 & 7.07 & 6.74 & - & 6.49 & \textbf{4.97} & 7.22 & 6.73 & - & 6.35 & \textbf{5.16} \\
BERT     & 7.18 & 6.76 & - & 6.37 & \textbf{5.83} & 7.35 & 6.70 & - & 6.66 & \textbf{6.02} \\
NCF      & 3.86 & 3.32 & 3.89 & 3.65 & \textbf{2.78} & 4.64 & \textbf{3.65} & 4.56 & 4.34 & 4.06 \\
DLRM     & 2.68 & \textbf{2.58} & 3.68 & 3.88 & 3.37 & 4.13 & \textbf{3.12} & 4.38 & 4.46 & 4.05 \\
VGG16    & 7.58 & 7.71 & 7.87 & 5.84 & \textbf{5.13} & 7.97 & 7.66 & 7.87 & 6.93 & \textbf{6.14} \\
  \bottomrule
  \end{tabular}}
  \vspace{-0.3cm}
\end{table*}

\subsection{Main Results}

\paragraph{Comparisons with Baseline Optimization Algorithms.}

We use a genetic algorithm (GA) in the literature \cite{krishnan2023archgym} as the baseline for the co-optimization problem. 
Additionally, we compare our approach with the state-of-the-art open-source optimization algorithm HASCO~\cite{xiao2021hasco},
which is a two-stage algorithm using multi-objective Bayesian optimization to explore the hardware design space and heuristic and Q-learning algorithms to optimize the mappings across all the layers of the network. Because the space defined in Figure~\ref{tab:space} is too large for HASCO, we limit it by using larger step sizes when discretizing the ranges of the number of \glspl{pe} and the buffer sizes. 

\begin{figure}[tb!]
  \centering
  \subfloat[Average latency] {\pgfplotsset{
    width =0.48\textwidth,
    height =0.22\textwidth,
}
\begin{tikzpicture}
\begin{axis}[
    ybar,
    y label style={at={(axis description cs:-0.08,0.5)},rotate=0,anchor=south},
    x label style={at={(axis description cs:0.5,-0.4)},rotate=0,anchor=north},
    bar width=4pt,
    enlargelimits=0.15,
    ylabel={\small Latency},
    symbolic x coords={GA, HASCO, w/o. rs., w/o. sc., \mname{}},
    xtick=data,
    ymin=1e5,
    ymode=log,
    log basis y=10,
    legend style={draw=none, at={(0, 1.02)}, anchor=south west, legend columns=2, font=\scriptsize},
    tick label style={font=\scriptsize},
]
\addplot coordinates {(GA,1.99e7) (HASCO,9.48e6) (w/o. rs.,9.24e6) (w/o. sc.,7.64e5) (\mname{},1.27e5)};
\addplot coordinates {(GA,1.49e7) (HASCO,1.24e7) (w/o. rs.,2.10e7) (w/o. sc.,1.69e6) (\mname{},2.01e5)};
\end{axis}
\end{tikzpicture}\label{fig:eff_t}} \hspace{0.3cm}
  \subfloat[Average LAS] { \hspace{-0.7cm} \pgfplotsset{
    width =0.48\textwidth,
    height =0.22\textwidth,
}
\begin{tikzpicture}
\begin{axis}[
    ybar,
    y label style={at={(axis description cs:-0.09,0.5)},rotate=0,anchor=south},
    x label style={at={(axis description cs:0.5,-0.4)},rotate=0,anchor=north},
    bar width=4pt,
    enlargelimits=0.15,
    ylabel={\small LAS},
    symbolic x coords={GA, HASCO, w/o. rs., w/o. sc., \mname{}},
    xtick=data,
    ymin=1e5,
    ymode=log,
    log basis y=10,
    legend style={draw=none, at={(1.05, 0.9)}, anchor=north west, font=\scriptsize},
    tick label style={font=\scriptsize},
]
\addplot coordinates {(GA,2.22e7) (HASCO,1.01e7) (w/o. rs.,1.02e7) (w/o. sc.,4.37e6) (\mname{},6.14e5)};
\addplot coordinates {(GA,2.5e7) (HASCO,1.03e7) (w/o. rs.,2.11e7) (w/o. sc.,4.15e6) (\mname{},6.85e5)};
\legend{Cloud, Edge}
\end{axis}
\end{tikzpicture}\label{fig:eff_tpa}}
  \caption{Average target metrics on different platforms for different methods.}
  \label{fig:avg-all}
\end{figure}
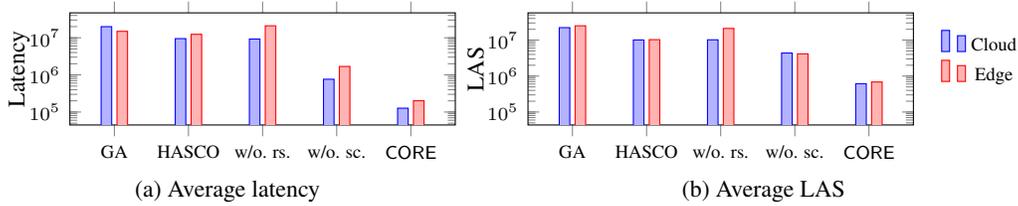

The results are shown in Table~\ref{tab:comp}, where we optimize the latency and the sum of latency and area (LAS) of the accelerator design for each \gls{dnn} model on cloud and edge platforms.
We optimize the entire model as a whole, and the reported values reflect the average reward computed over all layers. 
While HASCO achieves better metrics on smaller \glspl{dnn} like NCF and DLRM, \mname{} consistently obtains optimal results across other cases, demonstrating its superiority in terms of larger \glspl{dnn} and more complex design spaces.
We average the results across all models for each platform and illustrate in Figure~\ref{fig:avg-all} the average latency and LAS for each method.
\mname{} outperforms both GA and HASCO in terms of latency and LAS of the resulting design, achieving at least a $15 \times$ improvement in the optimal reward.

\paragraph{Ablation Study.}

We also conduct an ablation study to evaluate the effectiveness of the reward design and the graph-based scale-decoding strategy in \mname{}.
As detailed in Table~\ref{tab:comp}, the column ``w/o. rs.'' indicates the results when a negative penalty reward replaces the reward shaping upon constraint violations and anomalous designs, while the column ``w/o. sc.'' shows the outcomes when configurations are decoded independently from the produced actions, ignoring parameter dependencies.
The performance benefits are due to the structure-awareness and the mechanism to reason about constraints in \mname{}, which we expect to be transferable to other constrained domains.

The reward shaping can provide feedback when design points are invalid, enabling the agent to learn quantitative information, which facilitates the exploration process. 
In some cases with edge platforms, the algorithm without the reward shaping fails to find a valid design point throughout the exploration process, shown as ``-" in the table. 
Similarly, the dependency decoding strategy is instrumental in maintaining the feasibility of the decoded configurations, helping the agent understand the parameter dependencies and select feasible actions.
The ablation study shows a performance drop when either method is removed.

\paragraph{Sample Efficiency.}

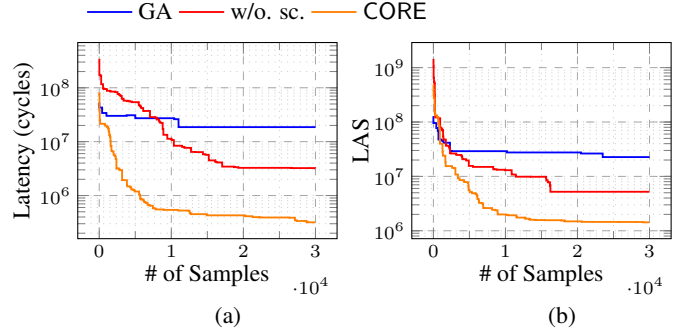
\begin{wrapfigure}{r}{0.64\textwidth}
  \centering
  \vspace{-0.6cm}
  \subfloat[] { \hspace{-0.3cm} \pgfplotsset{
    width =0.36\textwidth,
    height =0.3\textwidth,
}
\begin{tikzpicture}
\scriptsize
\begin{axis}[
minor tick num=1,
y label style={at={(axis description cs:-0.14,0.5)},rotate=0,anchor=south},
x label style={at={(axis description cs:0.5,-0.1)},rotate=0,anchor=north},
ylabel={Latency (cycles)},
xlabel={\# of Samples},
y label style={font=\small},
x label style={font=\small},
grid=both,
major grid style={dashed, line width=0.3pt, draw=gray!70},
minor grid style={dotted, line width=0.2pt, draw=gray!40},
ymode = log,
legend style={
	draw=none,
	at={(0, 1.05)},
	anchor=south west,
	legend columns=-1,
    font=\small
}
]
\addplot +[const plot mark right, blue, line width=0.7pt] [solid, mark=none] table [x={sample}, y={latency}] {
	sample 	latency
	16    53378224.0
    400  42909352.0
    1000 34040120.0
    3800 30130294.0
    5000 30906780.0
    10400 27217390.0
    11000 26115774.0
    18200 18615662.0
    30000 18615662.0
};
\addplot +[red, const plot mark right, line width=0.7pt] [solid, mark=none] table [x={sample}, y={latency}] {
	sample 	latency
16 	 343560384.0
112 	 172827424.0
272 	 169375136.0
528 	 116259928.0
592 	 97765272.0
688 	 95887640.0
1056 	 94790448.0
1552 	 87843720.0
2192 	 84922760.0
2544 	 83184600.0
2736 	 76549600.0
2928 	 74189808.0
3120 	 70655640.0
3376 	 62624080.0
3920 	 57640076.0
4208 	 56432420.0
4512 	 55560988.0
5456 	 54003480.0
5488 	 53056992.0
5616 	 47860936.0
5808 	 45271224.0
6096 	 42086080.0
6288 	 37180924.0
7040 	 37140032.0
7696 	 28543884.0
7968 	 28032814.0
8304 	 25753672.0
8464 	 23578522.0
8800 	 22342930.0
8896 	 17106654.0
9440 	 13412460.0
10048 	 11302432.0
10352 	 10736935.0
11488 	 8396710.0
11824 	 7997279.5
12704 	 7669882.5
12736 	 7477407.0
13312 	 6450466.5
14384 	 6400198.5
14736 	 5837012.0
15200 	 5706092.5
15216 	 4563651.0
16080 	 4489472.5
16128 	 4459980.0
16368 	 4167044.5
17056 	 4113254.5
18752 	 3443943.0
19296 	 3361201.75
26640 	 3266384.75
29952 	 3232624.5
30000 	 3194933.25
};
\addplot +[orange, const plot mark right,, line width=0.7pt] [solid, mark=none] table [x={sample}, y={latency}] {
	sample	latency
    16 82115120.0
    32 75767824.0
    64 37909972.0
    128 25277730.0
    800 21434284.0
    1056 19837174.0
    1184 18810072.0
    1280 18137792.0
    1312 17119208.0
    1392 15410158.0
    1424 13588028.0
    1456 13268755.0
    1520 12734207.0
    1552 12055613.0
    1584 10190891.0
    1696 8337880.5
    1776 6686014.5
    1984 6597470.0
    2144 5738918.0
    2288 5378360.0
    2384 4774454.5
    2432 3192751.5
    2784 3182386.75
    2848 3026076.5
    3152 2961633.5
    3168 2933087.5
    3184 1950374.75
    3840 1932033.375
    3872 1567900.625
    4288 1479520.125
    4448 1424453.875
    4752 1297201.5
    4912 1258721.875
    5536 1188739.0
    5568 1072820.375
    5680 1015915.75
    5712 913079.125
    5872 853106.875
    6112 847636.0
    6208 810944.375
    6416 763566.875
    6704 732349.6875
    6992 670814.5625
    7440 647680.125
    7808 579878.5
    8944 552430.0625
    10928 541273.0
    12416 527075.375
    12544 519683.0
    12720 470702.96875
    14912 451417.375
    20160 429271.46875
    20256 422612.78125
    21264 413226.1875
    22144 397704.75
    22352 393963.53125
    26416 391622.9375
    27152 387289.3125
    27392 347649.4375
    28576 337341.0
    28720 337001.5625
    28864 328437.15625
    30000 318440.75
};
\legend{GA, w/o. sc., \mname{}}
\end{axis}

\end{tikzpicture}\label{fig:eff_t}}
  \subfloat[] {  \hspace{-1.6cm}  \pgfplotsset{
    width =0.36\textwidth,
    height =0.3\textwidth,
}
\begin{tikzpicture}
\scriptsize
\begin{axis}[
minor tick num=1,
y label style={at={(axis description cs:-0.14,0.5)},rotate=0,anchor=south},
x label style={at={(axis description cs:0.5,-0.1)},rotate=0,anchor=north},
ylabel={LAS},
xlabel={\# of Samples},
y label style={font=\small},
x label style={font=\small},
grid=both,
major grid style={dashed, line width=0.3pt, draw=gray!70},
minor grid style={dotted, line width=0.2pt, draw=gray!40},
ymode = log,
legend style={
	draw=none,
	at={(-.4, 1.05)},
	anchor=south west,
	legend columns=-1,
    font=\small
}
]
\addplot +[const plot mark right, blue, line width=0.7pt] [solid, mark=none] table [x={sample}, y={latency}] {
	sample 	latency
0        128264720.0
400      95683032.0
600      77588968.0
700      69436496.0
1000     47714432.0
1100     47321168.0
1300     45785260.0
1400     44836096.0
2300     41596424.0
2500     29243428.0
10200    29088410.0
20500    27551292.0
23500    26352948.0
30000    22645972.0
};
\addplot +[const plot mark right, red, line width=0.7pt] [solid, mark=none] table [x={sample}, y={latency}] {
	sample 	latency
16       1449456256.0
32       1055620032.0
80       651679232.0
144      551260864.0
176      510717600.0
192      495759456.0
208      309714272.0
240      147832416.0
624      121552312.0
944      117339744.0
992      79569184.0
1168     76633760.0
1248     72588464.0
1344     70416896.0
1392     54445920.0
1728     47566808.0
1792     37297600.0
2144     36581648.0
2224     34572848.0
2304     32324346.0
2384     26790090.0
3184     26372582.0
3888     24777474.0
3968     23834360.0
4016     22700952.0
4544     20719972.0
4912     19632598.0
5488     15581774.0
8304     14947112.0
8448     13534453.0
8800     13475821.0
9936     13257482.0
10864    13007537.0
11440    10821158.0
13568    9868226.0
15632    9836967.0
15664    9604786.0
15872    8474012.0
16112    7809691.0
16176    7695838.0
16224    6777495.0
16240    6769669.5
25677    5215784.0
30000    5215784.0
};
\addplot +[const plot mark right, orange, line width=0.7pt] [solid, mark=none] table [x={sample}, y={latency}] {
	sample	latency
    16 	 486877504.0
32 	 390241824.0
48 	 314746784.0
160 	 276626272.0
176 	 208764032.0
240 	 152051984.0
496 	 132712328.0
672 	 124588096.0
736 	 91775864.0
768 	 90432000.0
848 	 85944832.0
896 	 73976712.0
912 	 44553016.0
1264 	 39976660.0
1312 	 28158212.0
1664 	 24035546.0
1696 	 18222776.0
2592 	 15445318.0
2896 	 14377376.0
3056 	 14122913.0
3408 	 10811068.0
3632 	 9248169.0
4256 	 8533288.0
4336 	 7980073.5
4768 	 7863578.5
4848 	 7596715.0
4880 	 6590964.0
5040 	 5579971.5
5344 	 5167794.5
5776 	 4897782.0
5936 	 4824447.5
6032 	 4679453.5
6112 	 4397663.5
6304 	 3817607.25
6320 	 3442765.5
6720 	 3256253.75
6912 	 3176493.0
7136 	 2621685.0
8144 	 2612754.75
8592 	 2431102.75
8912 	 2376010.25
10096 	 1996958.75
10384 	 1954184.75
11536 	 1925546.375
11552 	 1795616.75
12768 	 1748653.5
12816 	 1668703.0
13344 	 1666956.5
13360 	 1626519.125
14192 	 1593204.5
15392 	 1574348.875
17712 	 1563727.25
18176 	 1545188.25
18208 	 1536789.75
20592 	 1488134.75
22192 	 1457693.25
25648 	 1455486.375
25776 	 1454770.0
25872 	 1451761.625
28064 	 1450624.0
29072 	 1448442.125
30000 	 1429411.375
};
\end{axis}

\end{tikzpicture}\label{fig:eff_tpa}} 
  \caption{Optimized latency and LAS as the number of samples increases: (a) Change in optimized latency for ResNet50 with the cloud platform; (b) Change in optimized LAS for BERT with the edge platform.}
  \label{fig:tdf-all}
\end{wrapfigure}

The overall runtime can vary across different machines and parallelization settings, and it is mainly determined by the number of simulations required to achieve the optimized design. Therefore, we measure the number of sampled designs to achieve optimized latency and LAS to evaluate the space exploration efficiency.
The changes in latency for ResNet50 and a cloud platform and the changes in LAS for BERT with an edge platform can be seen in Fig.~\ref{fig:tdf-all}, where the x-axis represents the number of samples, and the y-axis denotes the value of the optimized objective. 
HASCO is not included in this comparison, as it is a two-stage algorithm that optimizes the hardware design space and mappings separately, which is not directly comparable to our single-step approach.
The method without reward shaping is also
omitted.
Both GA and the non-reward shaping method converge faster, but our work achieves convergence with significantly better metrics.
The results show that \mname{} requires fewer samples to achieve lower latency and LAS in both cases.

\paragraph{Discussion.}
{While our approach demonstrates strong performance across simulated metrics in efficiently navigating structured design spaces, its effectiveness depends on the fidelity of the underlying cost models.
In the context of \gls{dnn} accelerator design, our formulation further assumes static, compile-time mapping strategies. Extending the framework to support dynamic or runtime-adaptive dataflows presents a promising direction for future work.}
Interestingly, a recent method~\cite{guo2025deepseek} independently proposes a similar idea to compute the advantage via intra-batch reward comparisons in the context of optimizing LLM training dynamics, such as token sampling and update strategies.
This convergence highlights the generality of our approach, suggesting its applicability beyond \gls{dnn} accelerator design 
to other complex, constraint-rich domains such as compiler tuning, robotic system co-design, and automated system architecture search,
where simulation costs and structural constraints are similarly restrictive.
\section{Conclusion}
\label{sec:conclusion}

We presented \mname{}, a critic-free, one-step reinforcement learning approach for constraint-aware, simulation-guided design space exploration. \mname{} addresses critical limitations of existing methods by incorporating a surrogate objective based on relative advantages and enforcing feasibility constraints through a scaling-graph-based decoder. 
We validated \mname{}'s effectiveness on the challenging task of neural network accelerator hardware-mapping co-design, demonstrating significantly improved sampling efficiency and better optimization outcomes compared to state-of-the-art methods. 
Although demonstrated in the accelerator design domain, the \mname{} framework and its representations generalize broadly to other complex design tasks. Future work includes extending \mname{} to dynamic design environments, incorporating runtime adaptivity, and integrating it with symbolic reasoning or large language models for enhanced interpretability and scalability.

\setcitestyle{numbers}  

\bibliographystyle{unsrt} 




\appendix
\label{sec:appendix}
\section{Detailed Description of the Spatial \gls{dnn} Accelerator}

This appendix provides additional information on the spatial Deep Neural Network (DNN) accelerator co-design framework, facilitating understanding of hardware resource allocation, mapping strategies, and associated performance modeling.

\subsection{Spatial DNN Accelerator Resources}
\label{sec:dnn}

\subsubsection{Hardware Resources}

A spatial DNN accelerator primarily comprises an array of processing elements (PEs), each equipped with a multiply-accumulate (MAC) unit and Level-1 (L1) buffers for local data storage. An accelerator typically also includes a shared Level-2 (L2) buffer that acts as a bridge between off-chip memory and the on-chip L1 buffers, optimizing data fetching and reducing memory bandwidth demands.
Networks-on-Chip (NoCs) facilitate efficient operand distribution from the L2 buffer to individual PEs and the collection of partial outputs, subsequently stored back into the L2 buffer.

\subsubsection{Mapping Strategies}

Mapping strategies determine how computations of a DNN layer are partitioned and executed across hardware resources.

\paragraph{Tiling Strategy}
Tiling divides large tensors, such as weights and activations, into smaller blocks or "tiles" to fit within the L1 buffers of the PEs. Computation is conducted on one tile at a time, enhancing data reuse and minimizing redundant memory transfers. The choice of tile size impacts the amount of data that must be simultaneously stored within on-chip buffers.

\paragraph{Computation Order}
The sequence in which loop computations are executed significantly affects performance. For instance, the six-loop nest structure of 2D convolution operations (Fig.~\ref{fig:conv}) provides multiple possible orderings ($6!$ possibilities), each with distinct implications for data reuse and performance.

\paragraph{Parallelism Strategy}
This strategy specifies the number of PEs concurrently utilized and identifies which loop dimensions to parallelize at each memory hierarchy level, greatly influencing accelerator throughput and efficiency.
\begin{figure}[t!]
\centering
\includegraphics[width=0.45\textwidth]{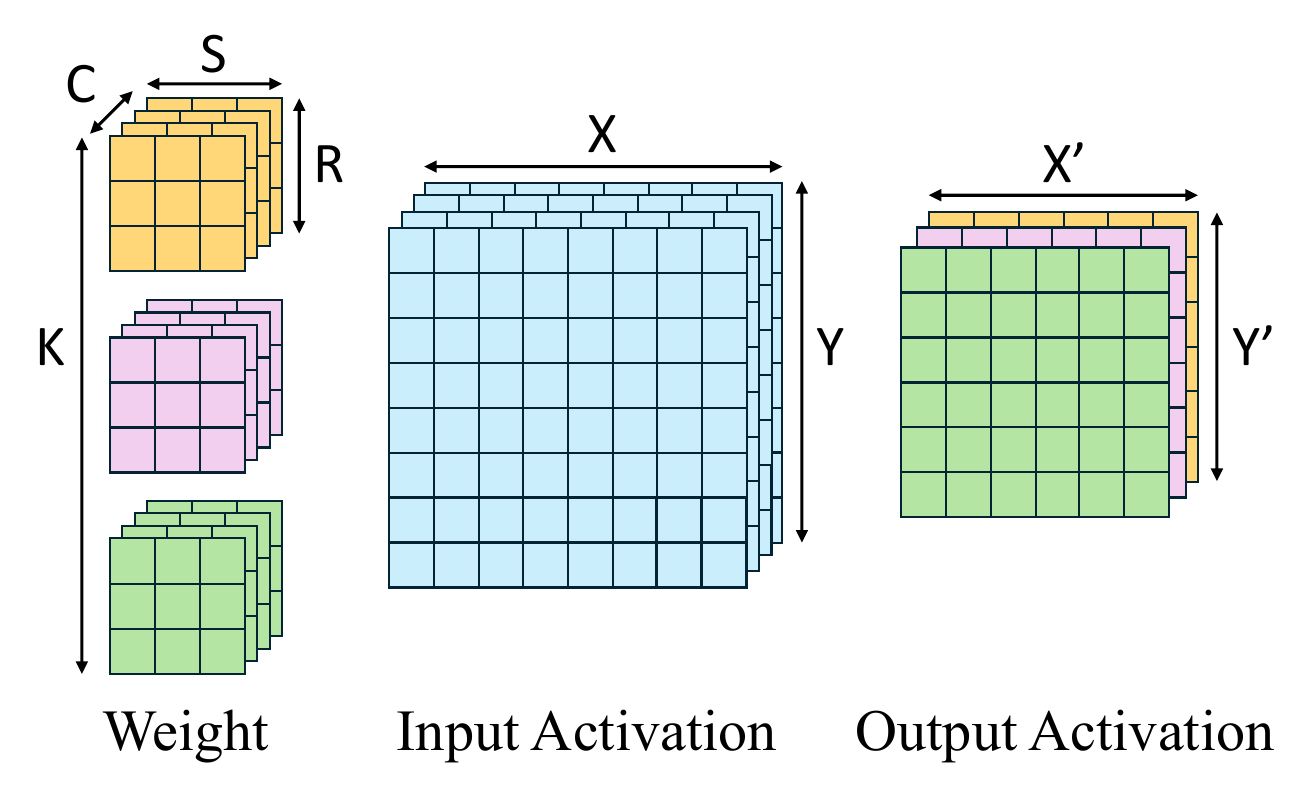}
\caption{Notation for the dimensions used in convolution operations. $\mathsf{K}$: output channels, $\mathsf{C}$: input channels, $\mathsf{Y}$: input height, $\mathsf{X}$: input width, $\mathsf{R}$: filter height, $\mathsf{S}$: filter width, $\mathsf{Y}'$: output height, $\mathsf{X}'$: output width.}
\label{fig:conv}
\end{figure}

\subsection{Accelerator Performance and Cost Modeling}

\subsubsection{DNN Model Characteristics}
Different DNN architectures (e.g., convolutional, fully connected, transformer layers) have unique computational patterns and data reuse opportunities. Convolutional layers dominate image-processing models, whereas fully connected layers appear frequently in language and recommendation models, each requiring tailored accelerator resources and mappings.

\subsubsection{Cost Models}
Accurate and efficient estimation of accelerator performance and resource usage is provided by analytical cost models such as MAESTRO~\cite{kwon2020maestro} and Timeloop~\cite{parashar2019timeloop}. These tools analyze hardware configurations and mapping strategies to quantify runtime, area, and power consumption, thus enabling informed decisions during accelerator design optimization.

\section{Policy Network Architecture}
\label{sec:nn}
In the experiment, we use a 4-layer multilayer perceptron with ReLU activations after each hidden layer. The input dimension is $512$. The hidden layers have widths $4096$, and the output layer matches the action space size. The layer dimensions are:
\[
512 \rightarrow 4096 \rightarrow 4096 \rightarrow 4096 \rightarrow \# \text{parameters for }\pi_\theta(a_1, ..., a_N; s_0).
\]
The network outputs conditional probability distributions for the design actions.
Specifically, each design parameter $p_i$ is modeled using a parametric distribution:
\begin{itemize}
    \item {Beta distribution}: requires {2 parameters} (e.g., $\alpha_i$, $\beta_i$)
    \item {Categorical distribution}: requires $k$ parameters for $k$ categories (e.g., logits)
\end{itemize}
Let $N$ be the number of design parameters, and let $\text{dim}(p_i)$ denote the number of parameters required for $p_i$. Then, the total output size of the policy network is:
\[
\# \text{parameters for }\pi_\theta(a_1, ..., a_N; s_0) = \sum_{i=1}^{N} \text{dim}(p_i)
\]



\section{Broader Impacts}

This work proposes a constraint-aware reinforcement learning framework to accelerate simulation-based \gls{dse} for \gls{dnn} accelerator co-design. The primary positive societal impact lies in reducing the computational and time cost of designing efficient AI hardware, which may enable deployment in resource-constrained domains such as edge computing, mobile healthcare, and environmental monitoring. Our method promotes sustainability by optimizing design with fewer simulations, contributing to greener hardware design workflows.

However, potential negative societal impacts include the risk of such optimization techniques being used to develop accelerators for ethically concerning applications. Furthermore, design automation could displace roles traditionally held by domain experts. These concerns warrant thoughtful deployment and governance when adopting such techniques.

We believe that transparent reporting, open-access implementation, and continued interdisciplinary dialogue are important to mitigate misuse and align technical advancement with societal values.

\end{document}